# Exploiting Qualitative Knowledge in the Learning of Conditional Probabilities of Bayesian Networks

**Frank Wittig and Anthony Jameson**
Department of Computer Science, University of Saarbrücken
P.O. Box 15 11 50, D-66041 Saarbrücken, Germany
{fwittig | jameson}@cs.uni-sb.de

## Abstract

Algorithms for learning the conditional probabilities of Bayesian networks with hidden variables typically operate within a high-dimensional search space and yield only locally optimal solutions. One way of limiting the search space and avoiding local optima is to impose qualitative constraints that are based on background knowledge concerning the domain. We present a method for integrating formal statements of qualitative constraints into two learning algorithms, APN and EM. In our experiments with synthetic data, this method yielded networks that satisfied the constraints almost perfectly. The accuracy of the learned networks was consistently superior to that of corresponding networks learned without constraints. The exploitation of qualitative constraints therefore appears to be a promising way to increase both the interpretability and the accuracy of learned Bayesian networks with known structure.

*If you don't know where you're going, you might wind up someplace else.* — Yogi Berra

## 1 INTRODUCTION

The following problem often arises when we use standard learning algorithms to learn the conditional probabilities of a Bayesian network (BN) with known structure and one or more hidden variables:[1] We have some fairly clear ideas about the qualitative relationships that must exist among the variables in the BN, and we are mainly interested in determining the quantitative relationships. But the learning algorithm uses no knowledge of qualitative relationships, so it "winds up someplace else": It produces a BN that may fit the data fairly well but that noticeably fails to exhibit the qualitative relationships that we expected. Such a network can be awkward to use and to explain to others, since it regularly violates natural expectations.[2] Moreover, we may suspect that a more accurate network could have been found which did fulfill our qualitative expectations.

Binder, Koller, Russell, and Kanazawa (1997, Section 7), after introducing the APN algorithm for learning BNs with hidden variables, proposed that it ought to be possible to guide the learning process by specifying qualitative constraints that the resulting network should satisfy. For example, a domain expert might state: "If the value of a variable $A$ increases, then the value of its child variable $B$ also increases". The purpose of the present paper is to work out and implement this proposal.

Our basic method for exploiting such constraints is to define a term within the network scoring function that reflects the overall extent to which qualitative constraints are violated by an intermediate learned network. If such a network rates poorly according to this criterion, the learning algorithm should tend to move on to alternative networks that rate better.

Another related approach, which goes a step further in the same direction as our approach, is to specify in advance specific types of functions describing the nature of the (uncertain) relationships between particular variables in the BN; and to apply learning techniques that are appropriate for these functions. For example, a linear relationship between a child and its parents could be learned using linear regression (see, e.g., Musick, 1996; Binder et al., 1997). But in many cases, it is not clear what specific functional relationship holds, even though it is clear that there must be a monotonic relationship of the sort mentioned above.

---

[1] For an overview of the general problem of learning Bayesian networks from data, see, e.g., Heckerman (1998).

[2] Some discrepancies may simply be due to the fact that the learning algorithm has in effect labeled the states of a variable differently than we expected, e.g., assigning to the state "On" the probabilities that we would expect for "Off", and vice-versa. These deviations from expectations may be easily correctible; but in practice they do not account for all of the violations of qualitative expectations.

The paper is organized as follows: After briefly reviewing the basic concepts and notation involved in the learning of Bayesian networks, we present a general conceptualization of how qualitative constraints can figure in the learning process. In Section 3 we then show how qualitative constraints can be suitably formalized. Section 4 shows how these ideas can be applied to two important learning algorithms for BNs, the APN and EM algorithms. Section 5 presents and discusses the results of tests of our approach in which synthetic data were used.

## 2 LEARNING BAYESIAN NETWORKS WITH KNOWN STRUCTURE

### 2.1 BAYESIAN NETWORKS

Formally, a BN $B = (G, \theta)$ consists of two components.[3] The first one is a directed acyclic graph $G$ that represents the causal independencies that hold in the domain to be modeled by $B$. Nodes represent random variables and directed links between nodes are commonly interpreted as causal influences between these variables. We restrict our attention in this article to BNs in which all of the variables are discrete.

BNs are characterized by the following independence assumption: Given the states of its parents, a node is independent of all its non-descendants in the BN. The second component of a BN is a vector $\theta$ of conditional probability tables (CPTs) $\theta_i$ that represent the (uncertain) relationships between nodes and their parents. A node's CPT consists of conditional probabilities for each state of the node conditioned on its parents' state configuration. A BN $B$ represents a joint probability distribution $P(X_1, \ldots, X_n)$ over the states of its variables $X = \{X_1, \ldots, X_n\}$. Exploiting the independence assumption of BNs, the joint probability distribution decomposes into a product of local conditional probabilities:

$$P(X_1, \ldots, X_n) = \prod_{i=1}^{n} P(X_i|pa(X_i)) = \prod_{i=1}^{n} \theta_i. \quad (1)$$

The term $pa(X_i)$ represents the set of all configurations of $X_i$'s parents, while $\theta_i$ is the CPT belonging to node $X_i$. Therefore, $\theta = (\theta_1, \ldots, \theta_n)$. $\theta_{ijk} = P(X_i = x_{ij}|pa(X_i) = pa_k(X_i)) = P(x_{ij}|pa_k(X_i))$ stands for the entry corresponding to the $j$th state of $X_i$ in $\theta_i$ when its parents take on their $k$th configuration $pa_k(X_i)$.

### 2.2 THE LEARNING PROBLEM

The general problem of learning the conditional probabilities $\theta$ of a BN $B$ with known structure can be formulated

---

[3]Detailed treatments of the theoretical and mathematical basis of BNs are given by, among others, Pearl (1988) and Castillo, Gutierrez, and Hadi (1997).

---

as follows: Given a data set $D = \{D_1, \ldots, D_s\}$ of $s$ *training cases* $D_i$, find a set of CPTs $\theta$ that optimizes a certain *scoring function* that describes the fitness of $B$ with respect to $D$. Every training case $D_i$ is an assignment of states for a subset of $B$'s nodes. A common scoring function is the *likelihood* $P(D|\theta)$ of $D$ with respect to $\theta$ for a given structure $G$. For computational convenience, the logarithm of this function is commonly used:

$$\ln P(D|\theta) = \sum_{i=1}^{s} \ln P(D_i|\theta). \quad (2)$$

The vector $\theta$ that maximizes this function (locally) represents a (locally) optimal set of CPTs, leaving out of consideration any prior qualitative knowledge about the CPTs. When a BN includes one or more hidden variables, it is in general infeasible to compute exact solutions to this problem (see, e.g., Heckerman, 1998). Therefore, approximative algorithms such as the ones discussed in Section 4 are used.

### 2.3 THE ROLE OF QUALITATIVE CONSTRAINTS

Suppose now that we have asked a domain expert whether the CPTs of the to-be-learned BN satisfy a particular set of qualitative constraints $C$ and that the expert has answered "Yes". How can we take this fact into account within the framework implied by Equation 2?

One conceptualization would be that the expert has hereby specified a prior probability distribution over the possible values of $\theta$. But within the maximum-likelihood framework, it is more appropriate to think in terms of the likelihood that the expert would answer "Yes" given various possible states of reality—i.e., various extents to which the constraints are really satisfied.

Concretely, suppose that we have defined a function $violation(\theta, C)$ that indexes the extent to which the CPTs $\theta$ violate the constraints $C$: $violation$ takes the value 0 if there is no violation at all and some positive value otherwise which increases with the seriousness of the violation.

Let us consider the likelihood that the expert answers "Yes" as a function of $violation(\theta, C)$. This likelihood should be equal (or close) to 1 if there is in reality no violation; and it should move toward 0 as the value of $violation(\theta, C)$ increases from its minimum of 0.

A computationally convenient function that meets these requirements is the following one:

$$P(answer = yes|\theta, C) = \exp(-w \cdot violation(\theta, C)). \quad (3)$$

Here, the positive weight $w$ determines how quickly the probability decreases from its maximum of 1 as the extent of constraint violations increases from its minimum of 0.

We can now view the expert's statement as a single—but especially significant—"observation" that can be taken into account along with the normal observations $D$ in the dataset. Accordingly, we can add the log-likelihood of this "observation" to the right-hand side of Equation 2 to obtain a modified likelihood of all observations:

$$\ln P(D|\theta) - w \cdot violation(\theta, C). \qquad (4)$$

The aim is now of course to find a (local) maximum for this log-likelihood. The term $violation(\theta, C)$ can be viewed as a penalty term which will cause the search algorithm to avoid solutions that violate the constraints, and the constant $w$ can be viewed as the weight of this penalty term.

The empirical results presented in Section 5 suggest that, when the constraints $C$ are in fact satisfied by the model that generates the data, a solution will typically be found for which the value of this term is (close to) 0.

It is clear that the function specified in Equation 3 is partly arbitrary. Indeed, determining the actual probabilistic relationship between constraint violations and expert judgments would require empirical research, and it might be impossible to find a useful domain-independent formulation. The above account can therefore be seen as specifying a possible scenario in which the penalty term in (4) can be given a probabilistic interpretation. Its intent is to clarify the relationship between the roles of the empirical data and the expert's judgment in guiding the search for a solution.

In order to be able to make use of (4), we have to answer two main questions:

1. How can the *violation* function best be defined and motivated for some useful class of constraints?
2. What algorithms can be used to find a (local) maximum of the scoring function of (4)?

The first question will be addressed in the next section and the second question in the subsequent sections.

## 3 FORMALIZING QUALITATIVE CONSTRAINTS

### 3.1 QUALITATIVE INFLUENCES

Within the framework of *qualitative probabilistic networks* (Wellman, 1990), Druzdzel and van der Gaag (1995) give formal probabilistic definitions of several types of qualitative relationships that can hold between nodes in a BN. The authors of the latter work employed these definitions in a method for combining different types of knowledge for the specification of the CPTs for BNs. They did not employ standard BN learning methods like the EM algorithm or gradient-based methods. Our method can be seen as an integration of parts of the method of Druzdzel and van der Gaag (1995) with standard BN learning algorithms. In this paper, we focus on the simple relationships that these authors call *qualitative influences*; but our method can be applied analogously to the more complex relationships that they also deal with.

The concept of a *qualitative influence* is only applicable if there is an ordering on the states of the nodes involved. Without loss of generality, we define $x_{i1} < x_{i2} < \ldots < x_{in_i}$ for every node $X_i$ with $n_i$ discrete states that is involved in a qualitative influence. A qualitative influence is denoted by $S^?(X_w, X_z)$, where $? \in \{+, -\}$ describes the quality ($+$ or $-$) of a monotonic relationship between a variable $X_w$ and one of its children $X_z$. Two kinds of qualitative influences exist: If a *positive* one holds, an increase in the state of $X_w$ causes an increase (or at least no decrease) in the state of $X_z$. If the relationship is negative, an increase in $X_w$'s state causes a decrease (or at least no increase) in $X_z$'s state.

Somewhat more formally, a positive qualitative influence $S^+(X_w, X_z)$ can be defined as follows: For any given value of $X_z$, an increase in the value of $X_w$ will not decrease the probability that the value of $X_z$ is equal to or greater than that given value.

Formally (cf. Druzdzel & van der Gaag, 1995): For all states $x_{zm}$ of $X_z$ with $m > 1$ and all distinct pairs of states $x_{wi}, x_{wj}$ of $X_w$ such that $i > j$ and for all possible state configurations $y$ of $X_z$'s parents other than $X_w$, the following inequality must hold:

$$P(X_z \geq x_{zm}|x_{wi}, y) \geq P(X_z \geq x_{zm}|x_{wj}, y). \qquad (5)$$

In terms of the conditional probabilities for individual states of $X_z$, this definition yields a set of inequalities of the following form:

$$\sum_{l=m}^{n_z} P(x_{zl}|x_{wi}, y) \geq \sum_{l=m}^{n_z} P(x_{zl}|x_{wj}, y). \qquad (6)$$

There exists one such inequality for each combination of an $x_{zm}$ such that $m > 1$, a pair $x_{wi}$ and $x_{wj}$ such that $i > j$, and a configuration $y$ of the states of $X_z$'s parents other than $X_w$.[4]

Negative qualitative influences are defined analogously.

---

[4] Actually, for the unambiguous specification of a constraint, we require only the inequalities that involve adjacent values of $X_w$, i.e., where $i = j + 1$, since the other inequalities are implied by the transitivity of the relation $\geq$. But in cases where a constraint has been violated, the redundant inequalities allow us to identify all of the values that are involved in the violation. As will become clear below, it is then possible to adjust all of these values simultaneously so as to eliminate the violation more quickly.

## 3.2 DEFINITION OF A VIOLATION TERM

We can now see how to define a suitable index of the overall extent to which a given set of CPTs $\theta$ violates a given set of constraints $C$, when these constraints concern qualitative influences—i.e., how to define a *violation* function of the type introduced in Section 2.3. Consider Inequality 6, which is part of the mathematical description of a positive qualitative influence of $X_w$ on $X_z$. We can write this inequality more generally as follows:

$$\underbrace{\sum_{l=m}^{n_z} P(x_{zl}|x_{wi}, y) - \sum_{l=m}^{n_z} P(x_{zl}|x_{wj}, y)}_{=:c'^{?wz}_{mijy}} \geq 0. \quad (7)$$

For every violated positive constraint, there has to exist at least one such inequality that is not satisfied—i.e., where the difference on the left-hand side of the equation is negative. Analogously, violations of negative constraints lead to values greater than 0.

A *partial violation term* corresponding to a single inequality can be defined as follows:

$$c^{?wz}_{mijy} := \begin{cases} -c'^{?wz}_{mijy} & \text{, if ?} = + \text{ and } c'^{?wz}_{mijy} < 0, \\ c'^{?wz}_{mijy} & \text{, if ?} = - \text{ and } c'^{?wz}_{mijy} > 0, \\ 0 & \text{, otherwise.} \end{cases} \quad (8)$$

The total violation term *violation*$(\theta, C)$ is defined as the sum of all of the relevant partial violation terms:

$$\text{violation}(\theta, C) := \sum_{m,i,j,y,w,z} c^{?wz}_{mijy}, \quad (9)$$

where ? stands for the quality (+ or −) of the constraint corresponding to $w$ and $z$. Note that, for each combination of variables corresponding to the indices $w$ and $z$, only one quality ? can exist, since it makes no sense to specify both a negative and a positive influence for these two variables.

## 4 USING CONSTRAINTS IN LEARNING ALGORITHMS

Having seen how to define the violation term required by (4), we can address the problem of finding a (local) maximum of the right-hand side of that equation. An analytic solution is not in general available, but various iterative search methods have been proposed. We will first discuss the use of the APN method (Binder et al., 1997; Russell, Binder, Koller, & Kanazawa, 1995), which can deal with the addition of the constraint violation term in a straightforward way. We will then turn to the EM method, with which dealing with the constraint violation term is less straightforward.

### 4.1 BASIC APN

The *adaptive probabilistic networks* method (APN) is a gradient-based algorithm for the learning problem formulated in Equation 2.

The computation of new values $\theta'$ for the CPT entries is done by taking (small) steps in the direction that is determined by the gradient $\nabla \ln P(D|\theta)$ of the log-likelihood function that constitutes the right-hand side of Equation 2:

$$\theta' = \theta + \alpha \nabla \ln P(D|\theta), \quad (10)$$

where $\alpha$ is a step-size parameter.

As was shown by Binder et al. (1997, Section 5.1), the partial derivatives of the log-likelihood function can be computed as follows:

$$\nabla^u_{ijk} \ln P(D|\theta) = \sum_{l=1}^{s} \frac{P(x_{ij}, pa_k(X_i)|D_l, \theta)}{\theta_{ijk}}, \quad (11)$$

where the superscript $u$ indicates that the gradient is still unprojected; that is, it has to be projected onto the constraint surface defined by $\sum_j \theta'_{ijk} = 1$, so that the new values $\theta'_{ijk}$ will continue to obey this fundamental constraint on the entries of any CPT. After the projection has been performed (as described by Binder et al., 1997, Section 4), the resulting gradient vector $\nabla \ln P(D|\theta)$ can be used in Equation 10. Binder et al. (1997, Section 5.3) and Russell et al. (1995, Section 7) present empirical results concerning the effectiveness of this method for learning the CPTs of BNs with hidden variables.

### 4.2 TAKING CONSTRAINTS INTO ACCOUNT WITH APN

To use APN with the extended scoring function of (4), we need to compute a slightly more complex gradient:

$$\nabla \ln P(D|\theta) - \nabla w \cdot \text{violation}(\theta, C). \quad (12)$$

The partial derivatives for the first term are of course the ones specified in Equation 11. For the second term, we can write:

$$\nabla^u_{ijk} w \cdot \text{violation}(\theta, C) = w \cdot v_{ijk}(\theta, C). \quad (13)$$

Each $v_{ijk}(\theta, C)$ is the partial derivative of the *violation* function with respect to the CPT entry $\theta_{ijk}$. This partial derivative is easy to compute, as can be seen from Inequality 7: Each partial violation term is a linear function of CPT entries, with each entry occurring at most once and having a coefficient of either +1 or −1. The only partial violation terms that contribute to the total violation term are the ones that correspond to inequalities that are not fulfilled at the current point $\theta$ in the search space.

It is straightforward to show that $v_{ijk}(\theta, C)$ can be computed as follows:

$$v_{ijk}(\theta, C) = v_{ijk}^-(\theta, C) - v_{ijk}^+(\theta, C), \quad (14)$$

where $v_{ijk}^-(\theta, C)$ is the number of unfulfilled inequalities which suggest that $\theta_{ijk}$ ought to be lower and $v_{ijk}^+(\theta, C)$ is the number of such inequalities that suggest that $\theta_{ijk}$ ought to be higher. It is therefore intuitively plausible, as well as mathematically sound, to add the gradient specified by Equation 13 to the one specified by Equation 11, as follows:

$$\nabla_{ijk}^u = \sum_{l=1}^s \frac{P(x_{ij}, pa_k(X_i)|D_l, \theta)}{\theta_{ijk}} - w \cdot v_{ijk}(\theta, C). \quad (15)$$

As with normal APN, it is still necessary to project this unprojected gradient onto the constraint surface defined by $\sum_j \theta'_{ijk} = 1$ before using it to compute the next step.[5]

We will now see how the log-likelihood functions in (2) and (4) can also be handled with the powerful and frequently used EM algorithm.

### 4.3 BASIC EM

The *Expectation Maximization* or *EM* algorithm (Dempster, Laird, & Rubin, 1977), when applied to our problem of maximizing the log-likelihood in Equation 2, proceeds as follows: After $\theta$ has been initialized to some vector of starting values, the algorithm performs two steps iteratively: The first step, called the *expectation-* or *E*-step, computes, for each $D_i$ in $D$, an expectation for the value(s) of the hidden variable(s) in $D_i$. (This computation involves, for each $D_i$, instantiating the observable nodes of a BN with CPTs corresponding to $\theta$ and evaluating this BN to derive a belief about each hidden variable.) The result of this step is a hypothetical dataset $D'$ which includes, in addition to the observed values of the observable variables, expectations concerning the hidden variables.

The second step, which is called the *maximization-* or *M*-step, computes new CPT values $\theta'$ which (locally) maximize the total log-likelihood of this hypothetical dataset—a task that is much easier than maximizing the log-likelihood of the real dataset (cf. Equation 2). These new values $\theta'$ always yield a log-likelihood of the real dataset that is at least as high as the one produced by the previous values $\theta$.

For our particular problem of learning the CPTs of BNs with hidden variables, the *E*-step and the *M*-step taken together yield a simple update rule (see, e.g., Castillo et al., 1997, p. 515):

$$\theta'_{ijk} = \frac{E_\theta[N_{ijk}]}{E_\theta[N_{ik}]}. \quad (16)$$

Here, $E_\theta[N_{ik}]$ is the expectation of the number of observations in the dataset in which the parents of the node $X_i$ take on their $k$th configuration $pa_k(X_i)$. $E_\theta[N_{ijk}]$ is the expectation of the number of such observations for which $X_i$ takes on its $j$th value $x_{ij}$. The latter expectation can be computed according to Equation 17:

$$E_\theta[N_{ijk}] = \sum_{l=1}^s P(x_{ij}, pa_k(X_i)|D_l, \theta), \quad (17)$$

while the former expectation is found through summation over $j$.

The update rule in Equation 16 is applied repeatedly until it converges on a (local) optimum for Equation 2, a solution which represents a maximum-likelihood estimate of the set of CPT entries.

### 4.4 TAKING CONSTRAINTS INTO ACCOUNT WITH EM

How can EM be used to maximize the extended scoring function (4) instead of the simple log-likelihood of the observed data? The most elegant approach would be to apply the expectation and maximization steps directly to the scoring function (4) so as to derive a modified update rule that could be used instead of (16). Unfortunately, the general EM approach is not equally easy to apply to all possible scoring functions; in particular, an attempt to apply it to the scoring function in (4) yields an interrelated set of nonlinear equations for which we did not find an analytic solution. The further pursuit of this approach therefore remains a matter for future research, which might, for example, consider the use of somewhat different scoring functions.

Because of the generally desirable properties of the EM approach, it seems worthwhile to pursue a theoretically less elegant modification of it which is capable of dealing with the extended scoring function of (4), albeit in a heuristic manner. Similarly, other researchers have developed variants of EM which are theoretically less justifiable than the pure form but which can be shown in practice to perform well, at least for some types of problems (see, e.g., Ortiz & Kaelbling, 1999; Bauer, Koller, & Singer, 1997). Like some of these methods, our approach combines EM with gradient ascent.

The basic idea is to alternate between two types of updates of the vector $\theta$ of CPT entries:

1. the standard EM update given by (16), which moves to an intermediate solution which yields a higher log-likelihood for the observations;

---

[5]In addition to the normalization mentioned above, these methods have to take into account that $\theta'_{ijk} \in [0, 1]$. This can be done by not allowing the learning algorithm to leave this search space. This may lead to situations where the learning procedures tend to follow the boundaries of the search space.

2. a gradient-based update which uses the gradient specified in Equation 13 and which aims only to decrease the extent of constraint violations that are found at the solution $\theta'$ that results from the EM update.

This results of applying this method are theoretically less predictable than the results for normal EM—or indeed for many of the variants of EM—, since the quality of the solution cannot be guaranteed to increase with each iteration: In principle, an EM update that increases the log-likelihood slightly can lead to a drastic decline in the extent to which the constraints are satisfied; and vice versa for a gradient-based update that slightly reduces constraint violations. On the other hand, if the specified constraints really do hold, the two goals of maximizing log-likelihood and maximizing constraint fulfillment are generally compatible; hence we would not expect the two types of update to work continually at cross-purposes.

The performance of this hybrid algorithm in practice will be investigated in the next section.

## 5 TEST OF THE METHOD

We conducted empirical tests using both of the two procedures described in the previous section. Since the initial results were qualitatively roughly similar, we conducted the most systematic tests for the modified EM procedure, because this procedure is more in need of empirical validation, given its partly heuristic nature.

Two network structures were used for the tests. Since theoretical interpretability is one of the motivations for the use of qualitative constraints, the first network structure comes from a domain in which the interpretability of a single hidden variable is important. The second—abstract—example demonstrates the feasibility of our approach for network structures involving more than one hidden variable.

### 5.1 EXAMPLE NETWORKS

Our first example BN, shown in the top part of Figure 1, could be used as a basis for an influence diagram for a hypothetical assistance system $S$ that presents sequences of spoken instructions to the user $\mathcal{U}$ (as, for example, a speech-based help system might do).[6]

An adaptation decision that $S$ has to make is whether to present a given set of instructions in a *stepwise* manner (i.e., $S$ presents its instructions one by one, allowing $\mathcal{U}$ to execute each one before presenting the next one) or in a *bundled* manner (all instructions are presented at once before $\mathcal{U}$ starts to execute the first one). One result of an experiment

[6]Explanations of the individual variables, along with the raw data from an experiment in this domain that involved 24 subjects, are available from http://w5.cs.uni-sb.de/~ready/. See also Jameson, Großmann-Hutter, March, and Rummer (2000).

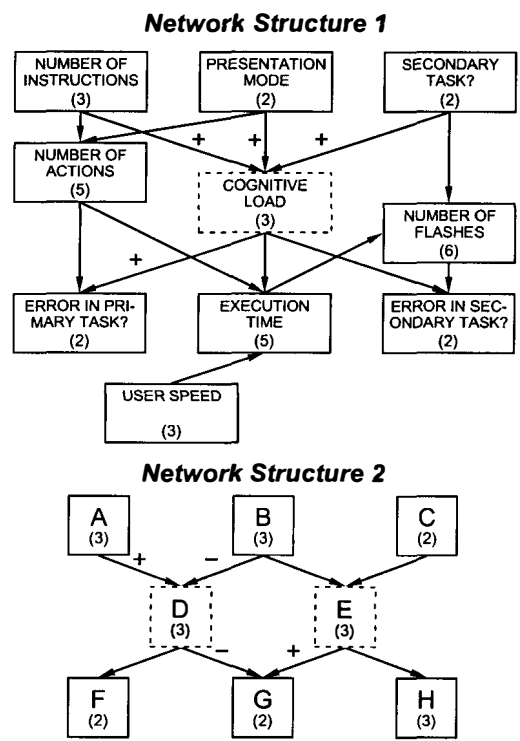

Figure 1. Structures of BNs Used for Testing.

(Nodes within dashed boxes correspond to hidden variables. For each of the arrows labeled + (or −), a positive (or negative) qualitative influence was postulated. The number in parentheses for each node is the number of states of that node.)

in this domain was that stepwise presentation reduced the number of errors $\mathcal{U}$ made but led to longer execution times.

We specified four qualitative influences involving the hidden variable COGNITIVE LOAD: Both psychological research and common sense led us to expect that the three parent variables of COGNITIVE LOAD would influence it positively. Moreover, higher COGNITIVE LOAD should increase the likelihood of ERROR IN PRIMARY TASK?.

The lower half of Figure 1 shows a second example BN. We offer no theoretical interpretation for it, but it enables us to test the effectiveness of the learning methods for network structures that include more than one hidden variable.

### 5.2 PROCEDURE

*Specification of original BNs.* We first manually specified a plausible BN for each structure just described, each of which satisfied the specified qualitative constraints. These *original BNs* were assumed for the rest of the evaluation to model the true causal relationships perfectly.

*Generation of synthetic learning data.* We then generated a sample of 1,000 learning cases using the first network structure and a sample of 200 cases with the second one. (Of course, the values generated for the hidden variables were not recorded.) We chose the numbers of 1,000 and 200 because (a) in the first domain a number of 1,000 seems to represent a realistic number of observations that one might be able to obtain and (b) we wanted to demonstrate the feasibility of the proposed method in the case of sparse data.

*Learning, starting with different initial BNs.* For each of the two structures, we generated ten BNs with randomly assigned initial values for the CPT entries. Each of these BNs was used as a starting point for two learning tasks: one using the standard EM algorithm and one using the extended algorithm that took the specified qualitative constraints into account.[7] In all cases, the learning procedure was terminated after 100 iterations.

*Evaluating the learned BNs.* To evaluate the accuracy of the learned BNs, we used the two original BNs to generate two sets of 10,000 and 5,000 test cases, respectively (again, without recording any values for the hidden variables). Each of the learned BNs was evaluated with respect to its *average negative log-likelihood per case*. In addition, the criterion of *average quadratic loss per case* was applied to the nodes ERROR IN PRIMARY TASK? and G in the first and second structures, respectively. (For both quality measures, lower values indicate better results.)

## 5.3 OVERALL RESULTS

The left-hand side of Figure 2 shows the main results for each of the 20 BNs that were learned for each network structure.

*Violation of constraints.* The narrow bars show that all of the BNs learned without constraints exhibited substantial violations even after 100 iterations. (The maximum possible values of *violation* were 63 and 54 for Network Structures 1 and 2, respectively.)[8] By contrast, the *violation* scores for the constrained BNs are mostly invisible in the graph because they are essentially zero.

*Fit to the test data (overall).* In the larger histogram for each network structure, the baseline shows the fit to the test data of the original BN that generated the test data. As one would expect, the fit of the learned BNs to the test data was worse than this baseline in every case; the fit is shown by the bars that project above the baseline. Looking at first only at these bars above the baseline and comparing the black bars with the gray ones, we see that the constrained BNs came closer to the baseline in 8 of the 10 cases with

---

[7] For the latter procedure, the violation weight $w$ (Equation 13) was set to 2.0. Moreover, the vector of $v_{ijk}$s used in the gradient step was rescaled so that the absolute size of its largest component was equal to the absolute size of the largest component of the preceding EM step.

[8] These quantitative values need to be interpreted with some caution, for the reason mentioned in Footnote 2.

Network Structure 1 (9% closer in terms of the average difference). With Network Structure 2, the constrained BNs were closer to the baseline in all 10 cases, the distance being 57% shorter on the average.

*Fit to the test data (selected variables).* With regard to the quadratic loss for the variables ERROR IN PRIMARY TASK? and G (not shown in Figure 2), the constrained BNs likewise fit the data consistently better than the unconstrained BNs. In fact, for the two network structures they come 50% and 66% closer to the relevant baseline, respectively, than the corresponding unconstrained BNs. Note that the criterion variables used here are directly involved in the specified constraints.

*Overfitting.* The fit of the learned BNs to the learning data indicates how much overfitting occurred during the learning process. Each of the learned BNs fit the learning data better than the original generating BN did. These results are shown by the bars that project below the baseline. We see that the unconstrained BNs overfit the learning data to a consistently greater extent than the constrained BNs. Thus, one advantage of using constraints seems to be that they offer a natural way to limit overfitting.

## 5.4 THE TIME COURSE OF LEARNING

To get a clearer picture of the reasons for the success of the learning procedure with constraints, we can examine the time course of the learning process for one typical BN for each network structure (see the right-hand side of Figure 2).

*Elimination of constraint violations.* In both cases, the procedure with constraints essentially eliminated the constraint violations within the first few iterations. (The values of the *violation* variable are not shown for the individual unconstrained BNs, since there was no typical pattern, aside from the fact that there were almost always substantial violations.)

*Evolution of the fit to the test data.* Looking at the two uppermost curves in each graph, we can distinguish two phases of the learning process. In the first phase, which typically lasts less than 10 iterations, the unconstrained BNs fit the test data better. Then, after a crossover point, the constrained BNs show a consistently better fit. This pattern can be understood in terms of the basic properties of the modified EM algorithm (Section 4.4): Initially, when there are substantial constraint violations, the normal EM steps alternate with gradient-based steps that serve solely to reduce constraint violations, perhaps at the expense of fit to the data. It is only after reaching a region of the search space in which the constraints are fulfilled that the algorithm can perform updates that are determined primarily by the goal of improving the fit to the data.

*Avoidance of overfitting as a key advantage.* With the two individual BNs shown on the right-hand side of Fig-

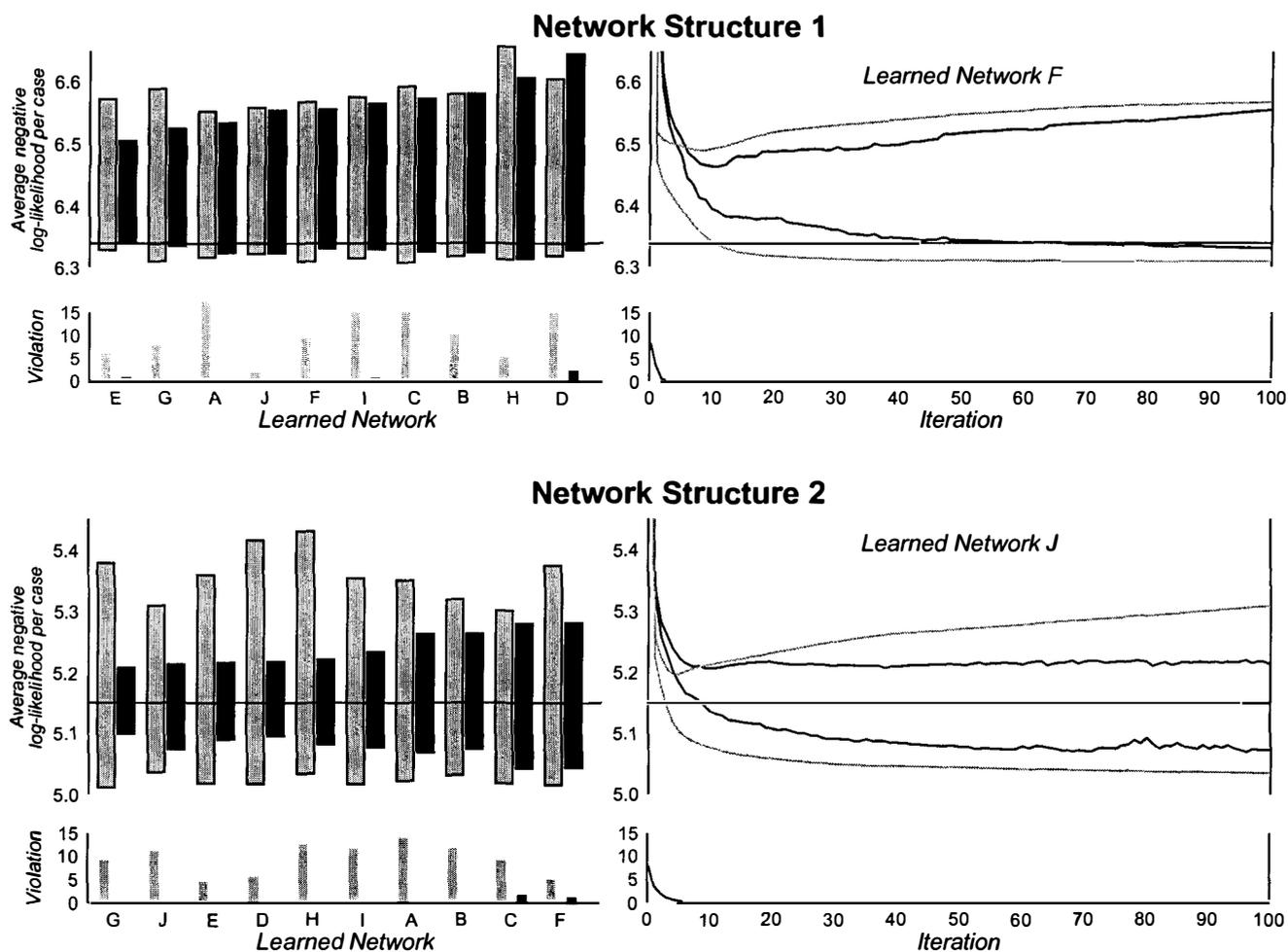

Figure 2. Results of the Tests With Synthetic Data.

(Black and gray bars and curves show results for BNs learned with and without constraints, respectively. The two uppermost learning curves in the graphs on the right show the fit to the test data, while the two curves below them show the fit to the learning data. The other aspects of the figure are explained in the text.)

ure 2, the optimal termination point for the learning procedure occurred somewhere between the 5th and the 10th iterations—whether constraints were used or not. In fact, with Network Structure 2, the unconstrained EM algorithm could have attained the best fit of all if it had known exactly where to terminate. On the other hand, terminating too early or too late without constraints could lead to significant loss of accuracy. In sum, the advantage of constraints in terms of fitting the data may not be that they yield a degree of fit that is unattainable without the specification of constraints. Rather, they appear to ensure a roughly equally good fit to the test data no matter when the learning process is terminated, as long as it is not terminated very early.

A certain amount of overfitting does seem to occur even with constraints, as is shown by the uppermost black curve for Network Structure 1. It will be interesting to investigate whether this overfitting mainly concerns the CPTs of variables for which no qualitative constraints have been specified.

## 6  CONCLUSIONS

While fitting test data is an important goal, it should be remembered that the elimination of constraint violations would in itself constitute an adequate motivation for the use of procedures such as the ones proposed here. Theoretical interpretability is a key goal in the application of BN learning techniques in many real-world systems that make use of BNs with hidden variables. Indeed, theoretical interpretability and explainability are important strengths of Bayesian networks generally.

Systems that use theoretically interpretable BNs may in some situations be better accepted by users. In particular, such a system's reasoning and decision making can be explained to users without the risk that the system will behave in a way that is incompatible with the explanation given. In the domain of our first network structure, we might formulate an explanation like "The presence of a secondary task increases the user's cognitive load; and higher cognitive load makes it more likely that the user will make an error."

Table 1. Summary of Contributions and Possible Extensions

| Contribution | Possible Extensions |
| --- | --- |
| A general conceptualization of the problem of incorporating prior qualitative knowledge into the process of searching for a Bayesian network that fits a given dataset of observations. | Application of this conceptualization to other types of learning problems. |
| A definition of a quantitative index of the extent to which a given BN violates specified qualitative constraints, based on work of Druzdzel and van der Gaag (1995). | Similar definitions for other types of qualitative constraints. |
| Specification and justification of adaptations of the basic APN and EM methods in accordance with the above contributions. | Derivation of a more theoretically justifiable adaptation of EM; adaptation of other search procedures, such as ELQ (Greiner, Grove, & Schuurmans, 1997). |
| Demonstration, using two quite different network structures and synthetic data, that the modified EM algorithm can learn BNs that fulfill the constraints (almost) perfectly while fitting the data better than the BNs learned by the unmodified algorithm. | Similar tests using larger networks and/or empirically collected data; systematic manipulation of parameters such as the step size, the weight of the violation term (Equation 4), and the proportion of nodes for which constraints are specified. |

Table 1 lists the contributions of the present paper and some corresponding possible extensions. Although there is clearly much that remains to be done, the results presented here seem to indicate that this work is worth doing.

### Acknowledgments

This research was supported by the German Science Foundation (DFG) in its Collaborative Research Center on Resource-Adaptive Cognitive Processes, SFB 378, Project B2 (READY). The implementations were done using HUGIN API from Hugin Expert A/S under a PhD license granted to the first author and with assistance from Björn Decker. We thank the anonymous reviewers for insightful comments.